\documentclass[10pt,twocolumn]{article}

\usepackage{amsmath}
\usepackage[a4paper,
            top=0.70in,
            bottom=0.75in,
            left=0.70in,
            right=0.70in,
            columnsep=0.25in]{geometry}
\usepackage{times}
\usepackage{microtype}
\usepackage{amsmath,amssymb,amsfonts}
\usepackage{booktabs}
\usepackage{graphicx}
\usepackage{adjustbox}
\usepackage{xcolor}
\usepackage[numbers,sort&compress]{natbib}
\usepackage[colorlinks=true,
            citecolor=blue,
            linkcolor=blue,
            urlcolor=blue]{hyperref}
\usepackage{enumitem}
\setlist{nosep}
\usepackage[utf8]{inputenc} 
\usepackage[T1]{fontenc}    
\usepackage{hyperref}       
\usepackage{url}            
\usepackage{booktabs}       
\usepackage{amsfonts}       
\usepackage{nicefrac}       
\usepackage{microtype}      
\usepackage{xcolor}         
\usepackage{booktabs}
\usepackage{adjustbox}
\usepackage{graphicx}
\usepackage{natbib}

\title{BERT Is Already a Denoiser: Cheaply Unlocking Iterative Generation in Masked Language Models}

\author{%
  Julio C. Amador D\'iaz L\'opez\thanks{
  SiftyML, LTD\\
  London, United Kingdom \\
  \texttt{julio@siftyml.com}}}

\begin{document}

\maketitle

\begin{abstract}

Masked language models (MLMs) such as BERT are conventionally treated as representation models rather than generators, despite being pretrained to reconstruct corrupted text. We ask whether this pretrained denoising
capability can be converted directly into iterative conditional generation
without explicitly training a diffusion model. We construct a simple decoder
that repeatedly applies a pretrained MLM to masked target positions and
progressively commits its highest-confidence predictions according to an
external cosine schedule. The model receives no diffusion timestep, learned
noise schedule, or reverse-diffusion objective.

Frozen BERT already exhibits substantial denoising capability, reconstructing
61.0\%, 37.5\%, and 24.4\% of masked tokens at 15\%, 45\%, and 60\%
corruption, respectively, although generation deteriorates sharply as visible
context disappears. This capability alone does not yield task-directed
generation: frozen BERT reaches only 11.0\% on ARC-Easy and 4.2\% on MMLU.
Ordinary MLM adaptation, however, rapidly converts the same architecture into
a usable generator. Full adaptation reaches 28.2\% and 24.6\%, respectively,
while parameter-efficient adaptation substantially reduces unmatched outputs.
On open-ended conditional generation, full MLM adaptation reaches
BLEU/ROUGE-L of $0.163/0.583$ on QQP and $0.107/0.327$ on Quasar-T, placing
it in the same broad performance range as published BERT models trained through diffusion 
without explicit diffusion training. Finally, applying the same generative
interface to stronger masked representations substantially improves
performance: RoBERTa-large reaches 44.2\% on ARC-Easy and 33.8\% on MMLU.

These results separate three components of masked iterative generation: pretraining supplies the representation and denoising operator, task adaptation determines what the model generates, and an iterative commitment rule supplies the decoding process. Explicit diffusion training is therefore not necessary to obtain effective iterative conditional generation from pretrained MLMs in the settings studied here.

\end{abstract}

\section{Introduction}
\label{sec:introduction}

Masked language models (MLMs) such as BERT~\citep{devlin2019bert} are typically treated as representation models rather than generators. Their pretraining objective reconstructs tokens removed from otherwise observed text, producing bidirectional representations that have proved effective across classification, retrieval, and structured prediction tasks\citep{DiazGarcia2025RelationExtraction}. In contrast, modern text generation is dominated by autoregressive models and, more recently, by diffusion-based approaches that formulate generation as iterative denoising.

The distinction is less clear than this division of labour suggests. A MLM is already trained to map corrupted language back toward clean language. It is then reasonable to assume that its pretrained denoising operator might already contain enough of the machinery that generation requires. If the model is repeatedly applied while its most confident predictions are progressively inserted into the context, masked-token prediction itself defines an iterative generation procedure.

This connection has motivated several approaches to generative masked
language modelling. BERT has been interpreted as a Markov random field from
which text can be sampled iteratively~\citep{wang2019bertmouth}, while
conditional masked language models have used iterative refinement for
non-autoregressive generation~\citep{ghazvininejad2019maskpredict}. More
recently, discrete diffusion models have formalized iterative text denoising
as a probabilistic generative process. Particularly relevant to our question,
DiffusionBERT initializes such a process from pretrained BERT, explicitly
exploiting the shared denoising structure of masked-language-model pretraining
and discrete diffusion~\citep{he2023diffusionbert}. This raises a more basic
question: \emph{how much additional generative machinery is required once a
pretrained masked language model already exists?}

We investigate this question by deliberately removing the diffusion machinery.
Starting from ordinary pretrained BERT-family MLMs, we introduce no forward
diffusion process, reverse diffusion parameterization, timestep embedding, or
learned noise schedule. Instead, generation consists of repeated MLM
evaluation followed by a simple external commitment schedule. At each
iteration, the model predicts all unresolved target positions, the
highest-confidence predictions are committed, and the remaining positions
stay masked for the next pass. Once committed, tokens are never remasked. The
result is a monotonic iterative decoder built directly around the pretrained
MLM.

This construction lets us separate three components that are often bundled
together in generative masked models: the \emph{representation and denoising
operator} learned during MLM pretraining, the \emph{task adaptation} that
aligns them with the desired output, and the \emph{decoding procedure} that
progressively turns masked predictions into a completed sequence. We study
these components through a sequence of experiments spanning frozen text
reconstruction, multiple-choice generation, paraphrase generation, question
generation, and different pretrained BERT-family representations.

We first ask whether useful iterative denoising is already present before any
generative adaptation. On corrupted natural text, frozen
\texttt{bert-base-uncased} reconstructs 61.0\% of masked tokens at 15\%
masking, 37.5\% at 45\%, and 24.4\% even when 60\% of the evaluated sequence
is masked. Distributional quality remains high at low corruption but
deteriorates sharply as visible context disappears, with repetition emerging
at very high masking ratios. These results establish both the capability and
its boundary: pretrained BERT is already an effective conditional denoiser,
but it does not become a useful unconditional generator simply by masking the
entire sequence.

We then test whether this denoising capability directly yields task-conditioned
generation. We find that it does not; when the question and answer choices are
visible but the answer itself must be generated through a masked target region,
frozen BERT reaches only 11.0\% accuracy on ARC-Easy and 4.2\% on MMLU with
most outputs failing to resolve to any candidate answer. However, ordinary
answer-only MLM adaptation changes this behavior rapidly. A parameter-efficient
BitFit condition raises ARC-Easy accuracy to 25.2\% while reducing unmatched
generations from 71.4\% to 12.8\%; full fine-tuning reaches 28.2\% on ARC-Easy
and 24.6\% on MMLU. In contrast, requiring the model to generate an intermediate
rationale increases the generative burden without improving final-answer
performance.

The same mechanism extends to open-ended conditional generation. On QQP, full MLM adaptation reaches BLEU $=0.163$ and ROUGE-L $=0.583$; on Quasar-T, it reaches BLEU $=0.107$ and ROUGE-L $=0.327$. These results place an ordinarily adapted MLM in the same broad performance range as a published explicitly trained masked-diffusion baseline on these tasks, despite requiring neither a diffusion objective nor diffusion-time conditioning~\citep{he2023diffusionbert}. The comparison is not a controlled head-to-head evaluation because the training and inference procedures differ. Rather, it provides evidence for the narrower claim tested here: explicit diffusion training is not required to obtain competitive iterative conditional generation from a pretrained MLM in these settings.

Finally, we ask what determines the competence of the resulting generator.
Frozen pseudo-log-likelihood controls reveal no large hidden
multiple-choice capability in BERT-base that is merely inaccessible to the
iterative decoder. Instead, replacing the underlying masked representation
changes generative performance substantially. BERT-large provides only a
modest gain over BERT-base, whereas RoBERTa-large reaches 44.2\% on ARC-Easy
and 33.8\% on MMLU under the same generative framework. Because BERT-large
and RoBERTa-large are similar in parameter scale, this comparison indicates
that generative competence depends strongly on the quality of the underlying
masked representation rather than model size alone.

Taken together, the results support a simple account of how pretrained MLMs
become generators. MLM pretraining supplies both the underlying representation
and a conditional denoising operator. Task-specific MLM adaptation aligns
these with the desired output, while an iterative commitment rule converts
repeated applications of the denoising operator into a decoding process.
The quality of the pretrained representation, in turn, constrains the
competence that the resulting generator can express. Our contributions are therefore fourfold:
\begin{itemize}
    \item We show that substantial iterative denoising is already present in
    frozen BERT and characterize how it degrades as visible context is
    progressively removed.

    \item We demonstrate that denoising capability and task-directed
    generation are empirically separable: frozen BERT is a useful denoiser
    but a poor task-conditioned generator, while ordinary MLM adaptation
    cheaply converts the same architecture into a usable generative
    interface.

    \item We show that standard MLM adaptation and simple iterative decoding
    can reach the performance range of an explicitly trained masked-diffusion
    model on conditional generation, without a diffusion objective, timestep
    conditioning, or diffusion sampling process.

    \item We show that the competence of the resulting generator tracks the
    underlying masked representation: stronger pretrained representations
    produce substantially stronger generators under essentially the same
    decoding procedure.
\end{itemize}

These findings narrow the role that must be assigned to explicit diffusion
training in masked text generation. They do not imply that diffusion provides
no additional benefits: probabilistic modelling, diversity, unconditional
generation, and alternative sampling behaviour remain distinct questions.
Rather, they show that the basic ingredients required for effective iterative
\emph{conditional} generation can already be latent in a pretrained masked
language model. In this regime, generation can be obtained by exposing and
adapting an existing denoising capability rather than by learning the
denoising process anew.

\section{Related Work}
\label{sec:related}

Our work sits at the intersection of three lines of research: masked language models as generative models, discrete diffusion for language, and recent attempts to simplify masked diffusion toward ordinary masked modelling.

\subsection{BERT and Masked Language Models as Generative Models}
\label{sec:related:bert_generation}

BERT~\citep{devlin2019bert} was introduced primarily as a bidirectional representation model trained with a masked language modelling objective. Although it is not autoregressive, its conditional token distributions can still define a generative procedure. Wang and Cho~\citep{wang2019bertmouth} formalized BERT as a Markov random field and showed that sentences can be sampled by repeatedly masking and resampling individual tokens. Their result established that masked language models are not intrinsically restricted to discriminative use.

Subsequent work has developed iterative generation methods based on masked prediction. Mask-Predict~\citep{ghazvininejad2019maskpredict} introduced confidence-guided iterative refinement for conditional sequence generation, while SUNDAE~\citep{savinov2021step} investigated text generation through repeated denoising. These approaches demonstrate that masked prediction can support non-autoregressive generation, although their training and decoding procedures differ from those of pretrained BERT. Our work instead examines how much of this capability is already present in an existing MLM and how much additional adaptation is required to exploit it.

More recently, Clavié et al.~\citep{clavie2025mask} introduced ModernBERT-Large-Instruct, a 0.4B-parameter encoder model that uses its MLM head for instruction-tuned generative classification. The model demonstrates strong zero-shot performance on classification and knowledge-based tasks, illustrating that pretrained masked representations can support task-directed outputs without a conventional classification head. Their approach focuses on single-token answers, whereas we investigate iterative multi-token generation.

\subsection{Discrete Diffusion Models for Language}
\label{sec:related:diffusion}

Diffusion models provide a probabilistic framework in which generation is defined through reversal of a progressively destructive forward process. While early diffusion models were developed primarily for continuous data, discrete variants extended the framework to categorical state spaces \citep{austin2023structured}. For language, discrete diffusion is attractive because corruption can be represented directly through token replacement or absorbing mask states.

DiffusionBERT~\citep{he2023diffusionbert} is the most direct precursor to our work. It combines pretrained BERT with an absorbing-state discrete diffusion process, motivated by the shared denoising structure of MLM pretraining and diffusion. The model learns a reverse diffusion process and investigates explicit timestep designs together with a token-frequency-dependent spindle noise schedule. For conditional generation, DiffusionBERT performs partial denoising conditioned on visible source text. Its results demonstrate that pretrained BERT provides an effective initialization for discrete diffusion and substantially accelerates convergence. 

Our work starts from the same empirical observation; i.e., that BERT is already a pretrained denoiser, but tests a different hypothesis. Instead of asking how best to adapt BERT into a diffusion model, we ask whether a diffusion model is necessary for the conditional-generation setting at all. We therefore remove the explicit forward process, reverse diffusion parameterization, timestep conditioning, and noise schedule, retaining only repeated MLM prediction and an external commitment schedule.

Other discrete diffusion approaches pursue more general generative modelling objectives. Score-Entropy Discrete Diffusion (SEDD), for example, learns discrete score ratios using a score-entropy objective and achieves competitive language modelling performance relative to autoregressive baselines \citep{lou2024sedd}. These approaches demonstrate that discrete diffusion can be a competitive generative paradigm in its own right. Our aim is narrower: we test whether pretrained masked language models already contain enough of the relevant denoising structure that a substantially simpler adaptation procedure suffices for conditional generation.

\subsection{Masked Diffusion and Classical MLM Objectives}
\label{sec:related:mdlm}

Recent work has further narrowed the conceptual gap between masked diffusion and classical masked language modelling. Sahoo et al.\citep{sahoo2024mdlm} derive a simplified masked diffusion objective whose training loss can be expressed as a mixture of classical MLM losses. Their Masked Diffusion Language Model (MDLM) achieves strong language-modelling performance and shows that encoder-only masked models can support efficient generative samplers. 

Related theoretical work argues that masked diffusion models can be time-agnostic: under suitable formulations, both training and sampling can be expressed without an explicit diffusion-time variable, making masked diffusion closely related to ordinary masked models
\citep{zheng2024timeagnostic}. 

These results are highly complementary to ours. They simplify the \emph{training objective} of newly trained masked diffusion models and show that diffusion-specific structure can collapse toward masked modelling. We instead begin from models that have already undergone large-scale MLM pretraining and ask whether that existing denoising representation can be reused directly. The distinction is therefore between \emph{training a generative masked model using an MLM-like objective} and \emph{retrofitting an already pretrained MLM into a conditional generator}. Our experiments target the latter.

\subsection{Positioning of the Present Work}
\label{sec:related:positioning}

The literature therefore establishes several pieces of the picture independently. BERT can be sampled iteratively \citep{wang2019bertmouth}; confidence-guided masked refinement can support conditional generation~\citep{ghazvininejad2019maskpredict}; dedicated denoising autoencoders can generate text~\citep{savinov2021step}; and discrete diffusion can combine effectively with pretrained masked models \citep{he2023diffusionbert}. More recent work further shows that masked diffusion objectives can simplify toward mixtures of MLM losses and may not require an explicit time variable~\citep{sahoo2024mdlm,zheng2024timeagnostic}.

Our contribution is to connect these observations through a direct empirical minimality test. We characterize the denoising capability already present in frozen BERT, show that this capability is distinct from task-directed generation, demonstrate that standard MLM adaptation plus iterative decoding is sufficient to acquire useful conditional generation, and show that the quality of the resulting generator tracks the competence of the underlying masked representation. In this sense, the question addressed here is not whether masked or diffusion models can generate language, but how much additional generative machinery is actually required once a pretrained MLM already exists.

\section{Methods}
\label{sec:methods}

We investigate whether pretrained masked language models (MLMs) can
perform iterative conditional generation through ordinary MLM adaptation
and a simple decoding procedure. Our approach combines a pretrained
MLM, which supplies conditional token predictions, with an iterative
decoder that progressively fills masked positions. Task-specific MLM
fine-tuning is used where adaptation is required.

We first characterize the denoising behavior of frozen BERT on
corrupted natural text, then examine its adaptation to multiple-choice
and open-ended conditional generation. Finally, we investigate how
generation performance depends on the underlying pretrained
representation.

\subsection{Iterative Masked Decoding}
\label{sec:methods:iterative}

Given a sequence $x^{(0)}$ containing a set of masked positions
$\mathcal{M}^{(0)}$, the MLM predicts a vocabulary distribution at each
masked position. At iteration $t$, the predicted token $\hat{x}_i^{(t)}$ and its confidence $c_i^{(t)}$
are:

\begin{equation}
\hat{x}_i^{(t)}
=
\operatorname*{arg\,max}_{v\in\mathcal{V}}
p_\theta(v\mid x^{(t-1)}),
\qquad
c_i^{(t)}
=
\max_{v\in\mathcal{V}}
p_\theta(v\mid x^{(t-1)}),
\label{eq:greedy_prediction}
\end{equation}

\noindent for every unresolved position $i\in\mathcal{M}^{(t-1)}$.

Rather than accepting all predictions simultaneously, the decoder
commits the $k_t$ positions with the highest confidence. Their
predicted tokens are inserted into the sequence, and the remaining
positions stay masked:

\begin{equation}
\mathcal{M}^{(t)}
=
\mathcal{M}^{(t-1)}
\setminus\mathcal{C}^{(t)},
\label{eq:remaining_masks}
\end{equation}

\noindent where $\mathcal{C}^{(t)}$ is the set of newly committed positions.

The MLM is then evaluated again on the partially completed sequence.
Committed tokens remain fixed, while unresolved positions are
reconsidered using the additional context. Generation terminates
when all masked positions have been resolved.

Unlike autoregressive decoding, the order of generation is not
constrained by sequence position. Confidence-based commitment can
resolve positions anywhere in the sequence, allowing subsequent
predictions to use context introduced on either side.

\paragraph{Commitment schedule.}
Let $N=|\mathcal{M}^{(0)}|$ be the initial number of masked positions
and $T$ the requested number of decoding passes, with
$T'=\max(1,\min(T,N))$. The number of positions remaining masked
after iteration $t$ follows a cosine schedule:

\begin{equation}
r_t
=
\operatorname{round}
\left[
N\cos\left(\frac{\pi t}{2T'}\right)
\right],
\qquad t=0,\ldots,T',
\label{eq:cosine_schedule}
\end{equation}

\noindent with $r_0=N$, $r_{T'}=0$, and monotonicity enforced by
$r_t\leftarrow\min(r_t,r_{t-1})$. The number of positions committed
at iteration $t$ is therefore $k_t=r_{t-1}-r_t$.

The schedule commits relatively few predictions initially and
progressively more as context accumulates. Iterations with
$k_t=0$ are skipped.

Unless otherwise stated, token selection is greedy (argmax),
corresponding to the zero-temperature limit. The first experiment
additionally evaluates alternative commitment orders and stochastic
token sampling as decoding controls.

\subsection{Reconstruction and Conditional Generation}
\label{sec:methods:interfaces}

The same decoder is used in two settings:

\paragraph{Full-sequence reconstruction.}
For denoising experiments, selected positions in a natural-language
sequence are replaced by the model's mask token:

\begin{equation}
x_i^{(0)}
=
\begin{cases}
[\mathrm{MASK}], & i\in\mathcal{M}^{(0)},\\
x_i, & i\notin\mathcal{M}^{(0)}.
\end{cases}
\label{eq:full_masking}
\end{equation}

The decoder reconstructs these positions while leaving unmasked
tokens unchanged. The primary configuration uses 16 requested
passes, confidence-based commitment, and greedy token selection,
without vocabulary restrictions.

\paragraph{Conditional target-region generation.}
For task-conditioned generation, the input consists of visible
conditioning context followed by a masked target region:

\begin{equation}
x^{(0)}
=
[\mathrm{context};
\underbrace{[\mathrm{MASK}],\ldots,
[\mathrm{MASK}]}_{N\ \mathrm{target\ slots}}].
\label{eq:conditional_masking}
\end{equation}

Only target positions are progressively resolved. The conditioning
context remains fixed, while committed target tokens become visible
on subsequent passes. The commitment schedule is calculated
independently for each example, allowing different target lengths
within a batch.

Multiple-choice generation uses four requested passes in the
primary configuration; open-ended generation uses eight.

\subsection{Task-Specific Decoding}
\label{sec:methods:task_generation}

\paragraph{Multiple-choice generation.}
For ARC-Easy and MMLU, the question and candidate answers are
provided as visible context. The primary prompt format is:

\begin{verbatim}
Question: <question>

Choices:
A. <choice0>
B. <choice1>
C. <choice2>
D. <choice3>

Answer: [MASK] ... [MASK].
\end{verbatim}

Inputs use the native special and mask tokens of each backbone,
with a maximum length of 512 tokens. When necessary, question
tokens are truncated before the choices and answer cue.

The primary configuration uses whole-word decoding and an
adaptive answer-length budget. If the reference answer contains
$w$ whitespace-delimited words, the target region contains $N=w+2$ mask slots. The additional slots provide a small generation buffer. Because this construction uses reference-answer length, we also evaluate fixed budgets of 4, 8, and 16 slots that do not depend on the reference answer.

Before token selection, model-specific special tokens and BERT's
\texttt{[unused*]} tokens are excluded. Under whole-word decoding,
BERT WordPiece continuation tokens beginning with \texttt{\#\#}
are also excluded. This restriction is not applied to RoBERTa's
BPE vocabulary. Excluded logits are set to $-\infty$ before
softmax.

\paragraph{Answer matching.}
Generated tokens are decoded to text and mapped to the visible
candidate answers. After lowercasing and removing non-alphanumeric
characters, exact matches are accepted directly. Otherwise,
candidates are scored using the larger of token-level F1 and
character-level sequence similarity. A candidate is accepted
only when its score is at least $0.5$ and uniquely exceeds the
alternatives.

Empty, ambiguous, and unmatched outputs are counted as incorrect.
We report both answer accuracy and the unmatched rate,

\begin{equation}
U=
\frac{1}{N_{\mathrm{eval}}}
\sum_{j=1}^{N_{\mathrm{eval}}}
\mathbf{1}
\left[
\hat{y}_j\not\mapsto\mathcal{A}_j
\right],
\label{eq:unmatched}
\end{equation}

\noindent where $\mathcal{A}_j$ is the candidate set for example $j$. No additional language model or learned matching component is used.

\paragraph{Open-ended generation.}
QQP paraphrase generation and Quasar-T question generation use
task-specific conditioning prompts followed by a fully masked
target region. Target length is fixed using the training
target-length distribution. Unlike multiple-choice generation,
subword tokens are permitted. Outputs are decoded using the
native tokenizer, with special tokens removed, and evaluated
directly against reference sequences.

\subsection{Models, Datasets, and Experimental Design}
\label{sec:methods:setup}

Our primary model is \texttt{bert-base-uncased}
\citep{devlin2019bert}. We subsequently evaluate BERT-large,
RoBERTa-large~\citep{liu2019roberta}, and a RoBERTa-large
backbone adapted to RACE~\citep{lai-etal-2017-race}.

We use four task families: WikiText~\citep{merity2017pointer}
for text reconstruction; ARC-Easy~\citep{clark2018arc} and
MMLU~\citep{hendrycks2021mmlu} for multiple-choice generation;
Quora Question Pairs (QQP) for paraphrase generation; and
Quasar-T~\citep{dhingra2017quasar} for question generation.

Unless otherwise specified, ARC-Easy and MMLU evaluations use
fixed seed-42 subsets of 500 examples, reused across the main
model and adaptation comparisons. 

\begin{table*}[t]
  \centering
  \caption{Summary of the experimental design.}
  \label{tab:experiments}
  \small
  \begin{adjustbox}{max width=\textwidth}
  \begin{tabular}{lllll}
    \toprule
    \textbf{Exp.}
      & \textbf{Model(s)}
      & \textbf{Data}
      & \textbf{Intervention}
      & \textbf{Primary purpose} \\
    \midrule

    A
      & BERT-base
      & WikiText
      & Frozen; 15--100\% masking
      & Pretrained denoising \\

    B
      & BERT-base
      & ARC-Easy, MMLU
      & Frozen iterative generation
      & Generation without adaptation \\

    C
      & BERT-base
      & ARC-Easy, MMLU
      & BitFit / Top-2 / Top-4 / Full
      & Generative adaptation \\


    D
      & BERT-base
      & QQP
      & MLM FT + iterative decoding
      & Paraphrase generation \\

    E
      & BERT-base
      & Quasar-T
      & MLM FT + iterative decoding
      & Question generation \\

    F
      & BERT-base
      & ARC-Easy, MMLU
      & Frozen PLL scoring
      & Discriminative competence \\

    G
      & BERT/RoBERTa variants
      & ARC-Easy
      & Frozen / BitFit / Full
      & Representation quality \\

    H
      & BERT/RoBERTa variants
      & MMLU
      & Full @100k
      & Representation transfer to MMLU \\

    \bottomrule
  \end{tabular}
  \end{adjustbox}
\end{table*}

\subsection{Experimental Conditions and Evaluation}
\label{sec:methods:experiments}

\paragraph{Frozen denoising (A).}
We evaluate frozen BERT-base on 256 WikiText blocks of 128 tokens,
using nested masking ratios of 15, 30, 45, 60, 75, 90, and 100\%.
The primary condition uses subword masking and 16 requested
decoding passes.

We additionally compare confidence-based commitment with random
and left-to-right commitment, holding the cosine schedule fixed.
A one-shot condition resolves all masked positions in a single
forward pass. Supplementary ablations vary decoding passes,
sampling temperature, and subword versus whole-word masking.

For stochastic decoding, tokens are sampled from
temperature-scaled vocabulary distributions,

\begin{equation}
\hat{x}_i^{(t)}
\sim
\operatorname{Categorical}
\left[
\operatorname{softmax}
\left(\frac{z_i^{(t)}}{\tau}\right)
\right],
\qquad \tau>0,
\label{eq:temperature_sampling}
\end{equation}

\noindent where $z_i^{(t)}$ denotes the MLM logits. The probability of the sampled token is used for confidence ranking.

We report masked-token reconstruction accuracy, GPT-2-large
negative log-likelihood (NLL), MAUVE,
and repeated four-gram rate (rep-4). These measure reconstruction,
external language-model predictability, distributional similarity,
and local repetition, respectively.

\paragraph{Frozen conditional generation (B).}
We apply the target-region decoder to frozen BERT-base on
ARC-Easy and MMLU. The primary condition uses four-pass
confidence decoding, adaptive whole-word answer budgets,
and the prompt format described above. Fixed-budget
conditions test sensitivity to reference-answer length.

\paragraph{MLM adaptation (C).}
We adapt BERT-base using masked-token prediction over the
answer region while keeping questions and candidate choices
visible. The objective is ordinary masked-token cross-entropy,

\begin{equation}
\mathcal{L}_{\mathrm{MLM}}
=
-\sum_{i\in\mathcal{M}}
\log p_\theta
\left(
x_i\mid x_{\setminus\mathcal{M}}
\right),
\label{eq:mlm_loss}
\end{equation}

\noindent where $\mathcal{M}$ denotes masked target positions. Training includes partially observed answer configurations, including high target corruption.

We compare BitFit, which adapts bias terms and the MLM output
components used by the implementation; Top-2 and Top-4, which
unfreeze the highest two or four transformer layers together
with the output components; and Full, which updates the
complete model. The principal comparison uses approximately
100k training examples.


\paragraph{Conditional text generation (D--E).}
We adapt BERT-base to QQP paraphrase generation and Quasar-T
question generation using masked-target prediction.
We evaluate frozen BERT, BitFit adaptation, and full
fine-tuning, reporting eight-pass decoding as the principal
configuration. Generation quality is measured using BLEU,
ROUGE-L, and BERTScore.

We also report published conditional-generation results
from DiffusionBERT~\citep{he2023diffusionbert} as a literature
reference. These results were obtained under different
training and inference configurations and do not constitute
a controlled head-to-head comparison.

\paragraph{Discriminative pseudo-likelihood (F).} We evaluate frozen BERT-base using normalized pseudo-log-likelihood (PLL) scoring \citep{salazar2020masked}. Pairwise and full-context formulations score candidate answers without free generation, providing a discriminative control on the same ARC-Easy and MMLU evaluation subsets.

\paragraph{Representation-quality experiments (G--H).} We repeat the generative interface with BERT-large, RoBERTa-large, and a RACE-adapted RoBERTa-large backbone. On ARC-Easy (G), we evaluate frozen generation, BitFit, and Full adaptation at 10k, 50k, and 100k examples, together with discriminative scores. For the RACE model, we additionally report its native multiple-choice performance.

Because the RACE-adapted backbone does not retain the standard pretrained MLM readout, we supply the original RoBERTa-large MLM head for iterative generation. Poor frozen generation in this configuration may therefore reflect a mismatch between the adapted backbone and the supplied head.

On MMLU (H), we evaluate the Full 100k checkpoints using the same 500-example, four-pass protocol. The BERT-base checkpoint is reused from experiment C. Table~\ref{tab:experiments} the experiments.

\subsection{Implementation}
\label{sec:methods:implementation}

At inference, the transformer processes the complete
partially observed sequence, but vocabulary logits are
computed only for masked positions relevant to the
current decoding step. This reduces memory consumption
without changing the conditional token predictions.
All decoding is performed in evaluation mode without
parameter updates.

The framework uses ordinary MLM training and an
external commitment schedule. No diffusion timestep,
noise-level embedding, learned reverse transition,
or additional generative model is introduced.
Once committed, tokens are never remasked.

\section{Results}
\label{sec:results}

\subsection{Pretrained BERT Supports Iterative Denoising}
\label{sec:results:denoising}

Frozen BERT-base reconstructs substantial portions of corrupted
natural text using iterative masked decoding. At 15\% masking,
61.0\% of masked tokens were recovered, with MAUVE $=0.899$
(Table~\ref{tab:denoising_mask_ratio}). Reconstruction accuracy
declined progressively as masking increased, reaching 37.5\%
at 45\% masking and 24.4\% at 60\%.

Distributional quality deteriorated more sharply than token
accuracy. MAUVE fell to 0.415 at 45\% masking and 0.265 at
60\%, while repetition increased rapidly above 60\%.
At 90--100\% masking, outputs increasingly collapsed into
repetitive sequences. Their lower GPT-2 NLL therefore reflects
greater predictability rather than improved generation quality.

\begin{table*}[t]
\centering
\caption{Frozen BERT-base reconstruction across masking ratios.
Results use 16-pass greedy confidence decoding. Lower NLL and
rep-4, and higher masked accuracy and MAUVE, are preferable.}
\label{tab:denoising_mask_ratio}
\small
\begin{tabular}{ccccc}
\toprule
Mask ratio & Masked accuracy & GPT-2 NLL & MAUVE & rep-4 \\
\midrule
15\%  & 0.610 & 3.75 & 0.899 & 0.011 \\
30\%  & 0.501 & 3.83 & 0.779 & 0.011 \\
45\%  & 0.375 & 3.92 & 0.415 & 0.010 \\
60\%  & 0.244 & 3.98 & 0.265 & 0.012 \\
75\%  & 0.126 & 3.80 & 0.168 & 0.067 \\
90\%  & 0.050 & 2.67 & 0.015 & 0.367 \\
100\% & 0.019 & 1.51 & 0.004 & 0.624 \\
\bottomrule
\end{tabular}
\end{table*}

At 45\% masking, confidence-based commitment achieved the
highest reconstruction accuracy (37.5\%), compared with
34.4\% for left-to-right and 33.6\% for random commitment
(Table~\ref{tab:denoising_strategy}). Random commitment,
however, obtained the highest MAUVE, indicating that
reconstruction fidelity and distributional similarity do
not necessarily favour the same decoding order.

\begin{table}[t]
\centering
\caption{Commitment-order ablation at 45\% masking.}
\label{tab:denoising_strategy}
\small
\setlength{\tabcolsep}{4pt}
\begin{tabular}{lccc}
\toprule
Strategy & Accuracy & MAUVE & rep-4 \\
\midrule
Confidence    & \textbf{0.375} & 0.415 & 0.010 \\
Left-to-right & 0.344 & 0.422 & \textbf{0.008} \\
Random        & 0.336 & \textbf{0.449} & \textbf{0.008} \\
\bottomrule
\end{tabular}
\end{table}

These results establish that pretrained BERT already
provides a useful conditional denoising operator, although
its ability to reconstruct text deteriorates substantially
when most of the conditioning context is removed.

\subsection{Task-Conditioned Generation Requires Adaptation}
\label{sec:results:frozen_generation}

The reconstruction capability of frozen BERT did not
translate directly into multiple-choice generation.
Accuracy reached only 11.0\% on ARC-Easy and 4.2\%
on MMLU, with unmatched-output rates of 71.4\%
and 80.8\%, respectively.

Ordinary answer-only MLM adaptation substantially
improved both measures (Table~\ref{tab:adaptation_strategies}).
After approximately 100k training examples, BitFit
increased ARC-Easy accuracy to 25.2\% and MMLU
accuracy to 17.4\%, while reducing unmatched outputs
to 12.8\% and 33.4\%. Full adaptation achieved the
strongest results, reaching 28.2\% and 24.6\%
accuracy, respectively.

\begin{table*}[t]
\centering
\caption{Multiple-choice generation before and after MLM
adaptation. All adapted models use approximately 100k
training examples and the same four-pass decoder.
Unmatched outputs are counted as incorrect.}
\label{tab:adaptation_strategies}
\small
\begin{tabular}{lcccc}
\toprule
& \multicolumn{2}{c}{ARC-Easy}
& \multicolumn{2}{c}{MMLU} \\
\cmidrule(lr){2-3}
\cmidrule(lr){4-5}
Adaptation & Accuracy $\uparrow$ & Unmatched $\downarrow$
           & Accuracy $\uparrow$ & Unmatched $\downarrow$ \\
\midrule
Frozen & 0.110 & 0.714 & 0.042 & 0.808 \\
BitFit & 0.252 & 0.128 & 0.174 & 0.334 \\
Top-2  & 0.242 & 0.226 & 0.176 & 0.454 \\
Top-4  & 0.258 & 0.122 & 0.174 & 0.328 \\
Full   & \textbf{0.282} & \textbf{0.088}
       & \textbf{0.246} & \textbf{0.158} \\
\bottomrule
\end{tabular}
\end{table*}

The reduction in unmatched outputs under BitFit is
particularly informative: limited parameter adaptation
was sufficient to make the pretrained model substantially
more effective at expressing task-directed answers
through the generative interface.



\subsection{Conditional Text Generation}
\label{sec:results:conditional_generation}

The same adaptation procedure extended to open-ended
conditional generation. On QQP, Full adaptation reached
BLEU $=0.163$, ROUGE-L $=0.583$, and BERTScore $=0.903$.
BitFit already achieved ROUGE-L $=0.470$ after 5k
training examples and 0.507 after 10k
(Table~\ref{tab:qqp}).

On Quasar-T, Full adaptation reached BLEU $=0.107$
and ROUGE-L $=0.327$. BitFit performed best at
10k examples and declined with further training
(Table~\ref{tab:quasar}).

\begin{table*}[t]
\centering
\caption{QQP paraphrase generation. DiffusionBERT results
are published reference values from
\citet{he2023diffusionbert}, not a controlled reproduction.
Training budgets count examples for our models.}
\label{tab:qqp}
\small
\begin{tabular}{lcccc}
\toprule
Model & Examples & BLEU $\uparrow$
      & ROUGE-L $\uparrow$ & BERTScore $\uparrow$ \\
\midrule
Frozen       & 0    & 0.004 & 0.101 & 0.666 \\
BitFit       & 5k   & 0.085 & 0.470 & 0.870 \\
BitFit       & 10k  & 0.107 & 0.507 & 0.883 \\
BitFit       & 100k & 0.121 & 0.533 & 0.889 \\
Full         & 100k & 0.163 & 0.583 & 0.903 \\
\midrule
DiffusionBERT (reported)
             & --   & 0.242 & 0.585 & -- \\
\bottomrule
\end{tabular}
\end{table*}

\begin{table*}[t]
\centering
\caption{Quasar-T question generation. DiffusionBERT
values are published reference results rather than
a controlled reproduction.}
\label{tab:quasar}
\small
\begin{tabular}{lcccc}
\toprule
Model & Examples & BLEU $\uparrow$
      & ROUGE-L $\uparrow$ & BERTScore $\uparrow$ \\
\midrule
Frozen       & 0    & 0.002 & 0.072 & 0.666 \\
BitFit       & 10k  & 0.050 & 0.233 & 0.744 \\
BitFit       & 100k & 0.026 & 0.190 & 0.741 \\
Full         & 100k & 0.107 & 0.327 & 0.781 \\
\midrule
DiffusionBERT (reported)
             & --   & 0.097 & 0.342 & -- \\
\bottomrule
\end{tabular}
\end{table*}

For context, the published DiffusionBERT results
report ROUGE-L $=0.585$ on QQP and $0.342$ on
Quasar-T. Our Full models reached 0.583 and
0.327, respectively, although the BLEU comparisons
differ across tasks. Because training and decoding
protocols are not matched, these values establish
similarity in the range of reported task performance,
rather than comparative superiority.

The central observation is that standard MLM
adaptation and iterative commitment can produce
substantial conditional-generation quality without
explicit diffusion-model training.

\subsection{Generation Performance Depends on the
Underlying Representation}
\label{sec:results:representation}

We next examined whether the relatively modest
multiple-choice accuracy of BERT-base reflected
limitations of the generative interface or the
underlying representation.

Frozen BERT-base achieved 31.4\% ARC-Easy accuracy
under pairwise pseudo-log-likelihood (PLL) scoring
and 25.6\% under full-context PLL. On MMLU, the
corresponding scores were 22.6\% and 23.8\%
(Table~\ref{tab:pll_control}). These results do not
reveal a large zero-shot discriminative advantage
that the generative procedure consistently fails
to recover, although frozen PLL and adapted
generation are not directly equivalent evaluations.

\begin{table}[t]
\centering
\caption{BERT-base discriminative PLL controls
and adapted generative accuracy (\%).}
\label{tab:pll_control}
\small
\setlength{\tabcolsep}{4pt}
\begin{tabular}{lcc}
\toprule
Method & ARC-Easy & MMLU \\
\midrule
Chance ($\sim$) & 25.0 & 25.0 \\
Full generation & 28.2 & 24.6 \\
PLL pairwise & \textbf{31.4} & 22.6 \\
PLL full-context & 25.6 & 23.8 \\
\bottomrule
\end{tabular}
\end{table}

Stronger pretrained representations yielded stronger
generators under the same iterative decoding procedure.
On ARC-Easy, Full adaptation reached 30.4\% for
BERT-large and 44.2\% for RoBERTa-large, compared
with 28.2\% for BERT-base
(Table~\ref{tab:backbone_arc}).

\begin{table*}[t]
\centering
\caption{ARC-Easy performance across pretrained
representations (\%). Discriminative scores use PLL
unless otherwise indicated; generative results use
four-pass confidence decoding.}
\label{tab:backbone_arc}
\small
\begin{tabular}{lccccc}
\toprule
Model & Discriminative & Frozen gen.
      & BitFit 100k & Full 100k & Full unmatched \\
\midrule
BERT-base
& 31.4 & 11.0 & 25.2 & 28.2 & 8.8 \\
BERT-large
& 29.6 & 11.8 & 28.2 & 30.4 & 9.8 \\
RoBERTa-large
& 41.2 & 17.0 & 34.2 & \textbf{44.2} & 5.2 \\
RoBERTa-RACE
& 33.8 & 0.4 & 35.4 & 41.8 & \textbf{3.6} \\
\bottomrule
\end{tabular}

\vspace{2pt}
\footnotesize
\emph{Note:} The RACE-adapted model also achieved
49.0\% using its native multiple-choice readout.
Its frozen generative condition uses a supplied
RoBERTa-large MLM head.
\end{table*}

The contrast between BERT-large and RoBERTa-large
is particularly informative. Despite similar
parameter counts, RoBERTa-large substantially
outperformed BERT-large, suggesting that
pretraining and representation quality contribute
beyond model scale alone.

The RACE-adapted backbone achieved 49.0\% with
its native multiple-choice readout but only
0.4\% in frozen iterative generation using a
supplied MLM head. After Full adaptation, its
generative accuracy increased to 41.8\%.
The frozen result therefore reflects, at least
in part, incompatibility between the adapted
representation and the supplied generative
readout.

The same ordering broadly persisted on MMLU.
RoBERTa-large reached 33.8\% accuracy and
RoBERTa-RACE 32.6\%, compared with 24.6\%
for BERT-base (Table~\ref{tab:mmlu_backbones}).

\begin{table}[t]
\centering
\caption{MMLU results after Full adaptation
at 100k training examples (\%).}
\label{tab:mmlu_backbones}
\small
\setlength{\tabcolsep}{4pt}
\begin{tabular}{lcc}
\toprule
Model & Accuracy & Unmatched \\
\midrule
BERT-base & 24.6 & 15.8 \\
RoBERTa-large & \textbf{33.8} & \textbf{8.6} \\
RoBERTa-RACE & 32.6 & 9.8 \\
\bottomrule
\end{tabular}
\end{table}

These findings indicate that the iterative
generation procedure does not impose a fixed
performance ceiling near the BERT-base results.
The competence of the resulting generator
depends substantially on the pretrained
representation and its compatibility with
the MLM readout.

\subsection{Summary}
\label{sec:results:summary}

Across the experiments, frozen BERT demonstrates
substantial iterative denoising capability, but
task-conditioned generation requires adaptation.
Ordinary MLM fine-tuning makes this capability
usable for multiple-choice and open-ended
generation, while stronger pretrained
representations produce stronger generators
under the same decoding procedure.

Together, the results support the view that
pretrained MLMs already supply the conditional
denoising operator needed for iterative
generation. Task adaptation aligns this
operator with the desired outputs, and
confidence-guided commitment provides a
simple mechanism for constructing complete
sequences.

\section{Discussion}
\label{sec:discussion}

Our results suggest that iterative conditional generation can be
decomposed into three components: a pretrained denoising operator,
task-specific adaptation, and a procedure for progressively
constructing the output. BERT already provides the first component
through masked-language-model pretraining. Ordinary MLM adaptation
can supply the task-specific behaviour, while confidence-guided
commitment turns repeated masked predictions into a completed
sequence.

The distinction is important because denoising and task-directed
generation are different capabilities. Frozen BERT reconstructs
partially corrupted natural text effectively, yet performs poorly
when asked to generate multiple-choice answers. The pretrained
model has learned to predict missing linguistic content, but
additional adaptation is needed to produce the particular outputs
required by a task.

\subsection{From Pretrained Denoising to Generative Behaviour}
\label{sec:discussion:adaptation}

The reconstruction experiments establish that useful iterative
denoising is already present in pretrained BERT. This capability
depends strongly on the available context: reconstruction remains
effective under moderate corruption but deteriorates as most of
the sequence becomes masked. Conditional generation provides a
different setting, in which a visible prompt or source sequence
can supply information about the masked target.

The adaptation experiments show that this distinction can be
exploited using the original MLM objective. Even restricted
parameter updates substantially improve multiple-choice
generation and reduce unmatched outputs. Full fine-tuning
provides further gains, but useful task-directed behaviour
emerges without retraining the complete model.

One interpretation is that adaptation serves two related
functions. \emph{Interface adaptation} makes task solutions
expressible through a masked output region, while
\emph{competence adaptation} modifies the representation
to improve the solutions available to that interface.
The BitFit results are consistent with the first mechanism,
whereas the additional gains from full fine-tuning suggest
a role for the second. These mechanisms are not separately
identified by the present experiments, but they provide
a useful framework for interpreting the different
adaptation regimes.

Under this interpretation, learning a conditional generator
does not necessarily require learning a new denoising
operator. A pretrained MLM may already supply that operator;
adaptation can instead establish the mapping between
conditioning information and the desired output.

\subsection{The Role of the Pretrained Representation}
\label{sec:discussion:representation}

The backbone comparisons indicate that the generative
interface does not impose a fixed performance ceiling.
BERT-large provides only modest improvements over
BERT-base, whereas RoBERTa-large produces substantially
stronger results under the same iterative decoding
procedure. Because BERT-large and RoBERTa-large have
similar parameter counts, model scale alone does not
explain the observed differences.

The discriminative controls provide complementary
evidence. Frozen BERT-base exhibits relatively limited
multiple-choice competence under pseudo-log-likelihood
scoring, while stronger pretrained representations
generally support stronger adapted generators.
Nevertheless, discriminative scoring and adapted
generation measure different capabilities, and their
results should not be interpreted as interchangeable.

The RACE-adapted RoBERTa experiment further illustrates
the importance of the output interface. Its native
multiple-choice readout performs substantially better
than frozen iterative generation using a supplied MLM
head. Generative adaptation subsequently recovers
considerable performance. Because the supplied head
was not jointly trained with the adapted backbone,
the initial failure cannot be attributed solely to
the representation.

Taken together, these findings suggest that generative
performance depends on both the information available
in the pretrained representation and the ability of
the MLM readout to express that information in the
required target format. Iterative decoding provides
the generation mechanism, but does not independently
determine the competence of the resulting model.

\subsection{Implications for Iterative Generation}
\label{sec:discussion:implications}

The conditional-generation results demonstrate that
ordinary MLM adaptation can support paraphrase and
question generation at a level broadly comparable
to published BERT adapted, masked-diffusion results. The comparison
with DiffusionBERT is not controlled, but it establishes
that an explicitly trained diffusion process is not
a prerequisite for obtaining this range of performance
on the evaluated tasks.

Our decoder provides a particularly simple construction.
At each iteration, the MLM predicts unresolved positions,
and the most confident predictions are committed according
to an external schedule. The model receives no diffusion
timestep or noise-level conditioning, and committed tokens
are never remasked. Generation therefore proceeds through
repeated conditional inference with an increasingly
complete context.

This construction does not establish that explicit
diffusion modelling is redundant. A learned diffusion
process may offer advantages in sampling diversity,
revision of earlier predictions, output-length modelling,
or generation from weakly conditioned inputs. These
possibilities are outside the present evaluation.
The contribution here is to isolate a simpler route
to iterative conditional generation using an existing
pretrained denoiser.

This has practical implications for the broader family
of masked language models. Many domains already possess
pretrained encoders developed from specialized corpora.
Where these models have learned useful conditional
representations, task-specific MLM adaptation and
iterative decoding may provide a route to generative
applications without training a new generative backbone
from scratch.

The broader architectural implication is that
representation learning, task adaptation, and generation
need not be treated as a single inseparable training
problem. Their contributions can be studied and
optimized independently, even though the resulting
capabilities remain interdependent.

\subsection{Limitations}
\label{sec:discussion:limitations}

Several limitations frame these findings. First, the experiments focus on conditional generation.
Frozen BERT deteriorates sharply at high masking ratios,
and the present results do not establish competitive
unconditional generation from a fully masked sequence.

Second, DiffusionBERT comparisons use published values
obtained under different training and inference
protocols. They indicate the range of achievable
conditional-generation quality but cannot establish
comparative efficiency or superiority. A controlled
comparison would require matched training data,
optimization budgets, decoding procedures, and
evaluation settings.

Third, the principal multiple-choice evaluations use
fixed subsets of 500 examples. Larger evaluations
and additional tasks are needed to establish more
precise estimates of absolute performance and
generalization.

Fourth, the primary multiple-choice decoder uses
reference-answer length to determine its output
budget. Fixed-length controls remove this dependency,
but a general-purpose generator should determine
output length independently. Open-ended generation
also relies on a predetermined target-length budget.

Finally, the experiments do not isolate which features
of MLM pretraining account for the observed denoising
capability, nor do they establish that confidence-based
commitment or the cosine schedule is optimal. The
present decoding procedure was chosen for simplicity
and to expose the generative capabilities of the
pretrained models.

\section{Conclusion}
\label{sec:conclusion}

Pretrained masked language models already contain a
useful conditional denoising operator. By combining
this capability with ordinary MLM adaptation and
confidence-guided iterative decoding, we obtain
task-conditioned generation without explicitly
training a diffusion model.

The experiments also establish an important boundary:
pretrained denoising alone does not guarantee useful
generation. Task adaptation is needed to direct the
model toward the required outputs, and the resulting
performance depends substantially on the quality
and accessibility of the underlying representation.

These findings identify a lightweight route from
pretrained masked representations to iterative
conditional generation, separating the acquisition
of denoising capability from the adaptation and
decoding mechanisms used to express it.

\bibliographystyle{plainnat}
\bibliography{references}


\end{document}